\documentclass[sigconf]{acmart}

\AtBeginDocument{%
  \providecommand\BibTeX{{%
    \normalfont B\kern-0.5em{\scshape i\kern-0.25em b}\kern-0.8em\TeX}}}

\usepackage{multirow}
\usepackage[utf8]{inputenc}
\usepackage[T1]{fontenc}
\usepackage{hyperref}
\usepackage{amsfonts}
\usepackage{graphicx}
\usepackage{amsthm}
\usepackage{amsmath}
\usepackage{subfigure}
\usepackage{multirow}
\usepackage{enumitem}
\usepackage{bbding}
\usepackage{colortbl}
\usepackage{arydshln}
\usepackage{utfsym}
\usepackage{arydshln}
\usepackage{tcolorbox} 

\theoremstyle{definition} \newtheorem{obs}{Observation}

\newcommand{\paratitle}[1]{\vspace{0.8ex}\noindent\textbf{#1}}

\setcopyright{acmlicensed}
\copyrightyear{2024}
\acmYear{2024}
\acmDOI{XXXXXXX.XXXXXXX}

\acmConference[MM'24]{the 32nd ACM International Conference on Multimedia}{October 28 - November 1,
  2024}{Melbourne, Australia.}
\acmISBN{978-1-4503-XXXX-X/18/06}

\begin{document}
\title{\textit{FakingRecipe}: Detecting Fake News on Short Video Platforms from the Perspective of Creative Process}

\author{Yuyan Bu}
\affiliation{%
  \institution{Institute of Computing Technology, Chinese Academy of Sciences}
  \institution{University of Chinese Academy of Sciences}
  \state{}
  \country{}}
\email{buyuyan22s@ict.ac.cn}

\author{Qiang Sheng}
\affiliation{%
  \institution{Institute of Computing Technology, Chinese Academy of Sciences}
  \state{}
  \country{}}
\email{shengqiang18z@ict.ac.cn}

\author{Juan Cao}
\affiliation{%
  \institution{Institute of Computing Technology, Chinese Academy of Sciences}
  \institution{University of Chinese Academy of Sciences}
  \state{}
  \country{}}
\email{caojuan@ict.ac.cn}

\author{Peng Qi}
\affiliation{%
  \institution{National University of Singapore}
  \state{}
  \country{}}
\email{peng.qi@nus.edu.sg}

\author{Danding Wang}
\affiliation{%
  \institution{Institute of Computing Technology, Chinese Academy of Sciences}
  \state{}
  \country{}}
\email{wangdanding@ict.ac.cn}

\author{Jintao Li}
\affiliation{%
  \institution{Institute of Computing Technology, Chinese Academy of Sciences}
  \state{}
  \country{}}
\email{jtli@ict.ac.cn}
\renewcommand{\shortauthors}{Bu et al.}

\begin{abstract}
As short-form video-sharing platforms become a significant channel for news consumption, fake news in short videos has emerged as a serious threat in the online information ecosystem, making developing detection methods for this new scenario an urgent need. Compared with that in text and image formats, fake news on short video platforms contains rich but heterogeneous information in various modalities, posing a challenge to effective feature utilization. Unlike existing works mostly focusing on analyzing \textit{what is presented}, we introduce a novel perspective that considers \textit{how it might be created}. Through the lens of the creative process behind news video production, our empirical analysis uncovers the unique characteristics of fake news videos in material selection and editing. Based on the obtained insights, we design \textbf{FakingRecipe}, a creative process-aware model for detecting fake news short videos. 
It captures the fake news preferences in material selection from sentimental and semantic aspects and considers the traits of material editing from spatial and temporal aspects. To improve evaluation comprehensiveness, we first construct FakeTT, an English dataset for this task, and conduct experiments on both FakeTT and the existing Chinese FakeSV dataset. The results show FakingRecipe's superiority in detecting fake news on short video platforms.
\end{abstract}



\maketitle

\section{Introduction}

\begin{figure}[t]  
    \centering    
    \includegraphics[width=1\linewidth]{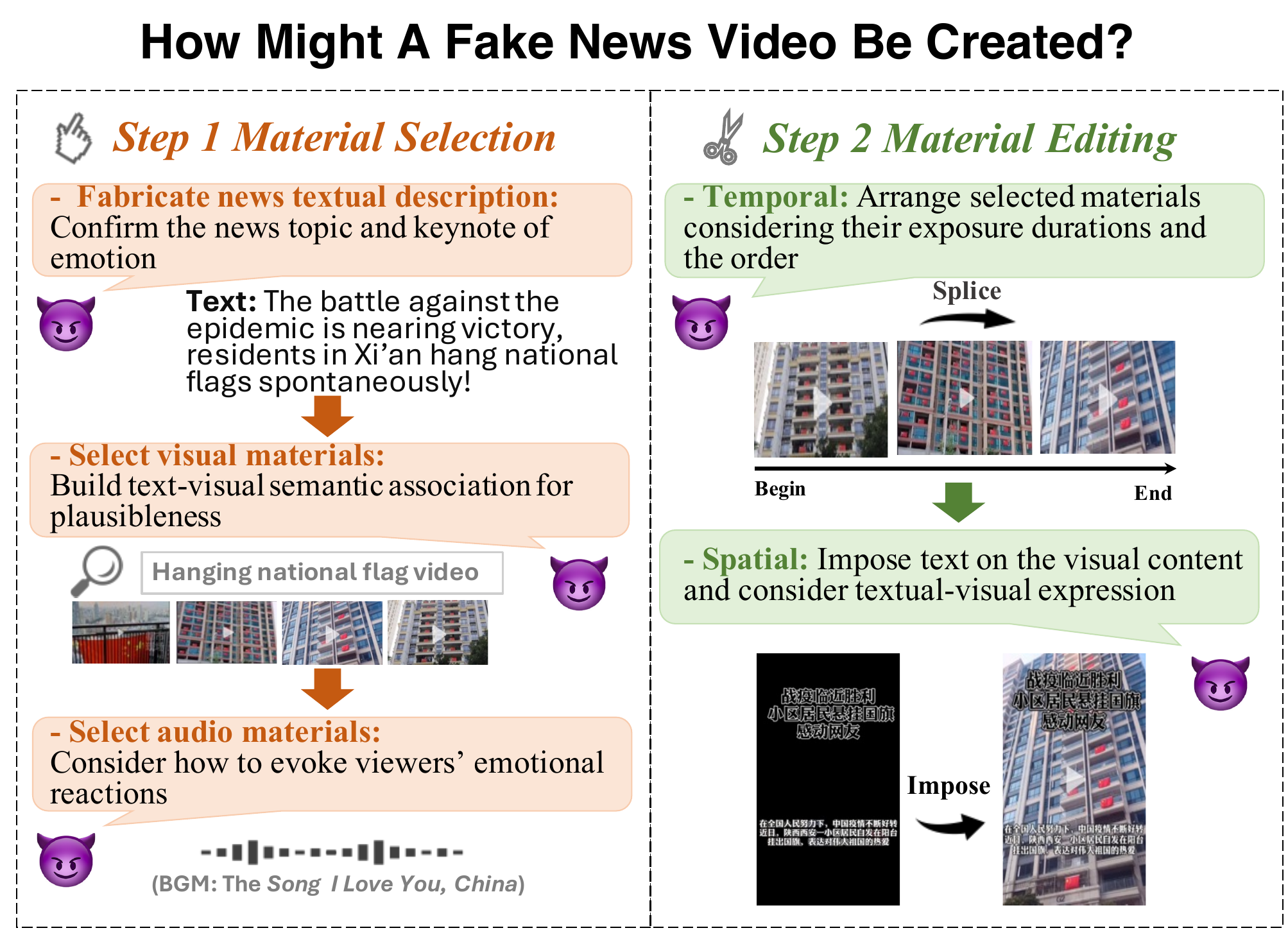}
    \caption{A fake news video about residents hanging national flags amid the COVID-19 pandemic in China, exhibited along with the speculated creative process. The text was translated into English.}
    \label{fig:videopost} 
    \vspace{-1em}
\end{figure}

In recent years, short-form video-sharing platforms like TikTok have been increasingly popular on mobile Internet and revolutionizing how people consume news~\cite{niu2023building,hendrickx2023newspapers}. According to Pew Research Center, by 2023, 33\% of U.S. adults have ever used TikTok~\cite{USASoicalMediaUse}, with nearly 43\% of these users frequently sourcing their news from this platform~\cite{TikTokNewsConsumption}. However, the prevalence of news consumption on short video platforms also encourages the emergence and spread of fake news videos, posing new serious threats to the online information ecosystem~\cite{sundar2021,bu2023combating}. Consequently, customizing methods for detecting short video fake news is of urgent need.

Unlike fake news in text or image formats, fake news on short video platforms shows unique characteristics and is increasingly indistinguishable from real news, posing new challenges to developing effective detectors~\cite{bu2023combating}. First, the easy-to-use video editing tools largely democratize news creation and enable almost everyone to edit a news video on par with professional journalists~\cite{niu2023survey}, making the edit traces widely exist in both real and fake news videos. Second, due to the public nature of short video platforms, even a real news video is likely to be repurposed or re-edited for news faking. 
However, existing methods for fake news video detection mostly follow ideas from the research line of text-image-based detection and focus on modeling \textit{what is presented} via analyzing the authenticity of multimodal content (\textit{e.g.}, detecting \textit{deepfakes}~\cite{ganti2022}) and modeling cross-modal correlation~\cite{qi2022fakesv,choi2021cikm,shang2021tiktec}, which are more likely to be misled by edited and repurposed contents.
Faced with the more vague boundary between the real and the fake, it is necessary to find new perspectives and capture more effective clues for fake news video detection.

\textbf{In this paper, we propose to switch the perspective from analyzing \textit{what is presented in a fake news video} to considering \textit{how it might be created}.} Our idea is based on a straightforward assumption: Fake news creators on short video platforms often lack first-hand, genuine news materials and professional editing skills while aiming to produce fake news for specific purposes intentionally~\cite{shu2017fake}. 
This might leave unique characteristics of the resulting video. \figurename~\ref{fig:videopost} provides an intuitive example of the creative process of a fake news video about hanging national flags during the COVID-19 pandemic. 
The process typically unfolds in two main phases: \textbf{material selection} and \textbf{material editing}.
For selecting the material, the creator first confirmed the news topic (i.e., residents hanging national flags during the pandemic) and the positive sentiment keynote and crafted an attractive narrative that diverges from the truth.
Due to the lack of real visual materials (unlike real news), the creator had to repurpose historical materials collected from the Internet to make the fake video more convincing.
Finally, an emotionally charged song is selected to impress audiences and achieve its underlying purpose.
For the editing phase, the creator might consider arranging materials from the temporal and spatial views with the help of simple editing techniques. Due to the constraint of material sufficiency and editing skills, the collected visual materials were arranged with simple splicing temporally, and the text material was then spatially overlaid on the visual content for a straightforward textual-visual expression.
Through this example, we intuitively find that the production of fake news videos may leave the nuances different from that of real ones in terms of material selection and editing. Therefore, modeling from the creative process perspective may help us capture more valuable instrumental clues for fake news video detection.

Inspired by the observation, in this paper, we first quantitatively examine how effective the clues from the creative process perspective are in distinguishing fake and real news videos via an empirical analysis (Section~\ref{sec: empirical_analysis}). The results validate that statistical discrepancies exist between real and fake news videos in material selection and editing. For instance, we find that compared with real ones, fake news videos exhibit a propensity for selecting more emotionally charged music, using a limited palette of colors, and adopting a less dynamic on-screen text presentation.
Based on the empirical analysis, we design \textbf{FakingRecipe}, a creative process-aware model for detecting fake news short videos.\footnote{The creative process of faking a news video is metaphorically similar to cooking a dish following a recipe, so we use \textbf{FakingRecipe} to highlight the model's uniqueness.} FakingRecipe is a dual-branch network that models the characteristics of material selection and editing. In the two branches, the Material Selection-Aware Modeling (MSAM) module extracts multimodal features via attention to capture the sentiment resonance between audio and text and the semantic relevance between text and visual frames. The Material Editing-Aware Modeling (MEAM) module models typical spatial and temporal editing behaviors, via 1) analyzing the visual area and on-screen texts for the spatial editing; and 2) building hierarchical temporal structure that considers both intra- and inter-segment fusion for temporal editing.
Ultimately, predictions from both branches are integrated through a late fusion function for the final prediction. Experiments on two real-world datasets demonstrate the superiority of the proposed FakingRecipe over seven baseline methods.
Our main contributions are as follows:

\begin{itemize} [leftmargin=0.5cm]
\item \textbf{Idea:} We for the first time consider the creative process as a pivotal aspect for detecting fake news on short video platforms and demonstrate the feasibility of this perspective through empirical analysis. 

\item  \textbf{Method:} We propose FakingRecipe, a novel dual-branch model for fake news video detection that captures useful clues from the perspective of the creative process, \textit{i.e.}, material selection and editing phases.

\item \textbf{Resource \& Effectiveness:} We construct FakeTT, an English short video dataset for fake news detection. Extensive experiments on both FakeTT and the public Chinese FakeSV dataset show the superiority of FakingRecipe over existing methods in fake news video detection. We will publicly release the new dataset to facilitate further research\footnote{\url{https://github.com/ICTMCG/FakingRecipe}}.

\end{itemize}

\begin{figure}[t]
    \centering
    \begin{minipage}{0.48\linewidth}
        \centering
        \vspace{-0.3cm}
        \includegraphics[width=\linewidth]{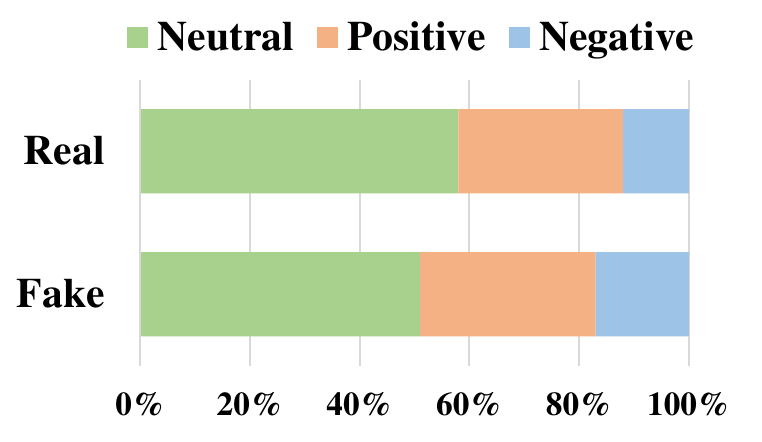}
        \vspace{-0.4cm}
        \caption{Sentiment analysis of audio material.}
        \label{fig:AudioEmo_fakesv}
    \end{minipage}%
    \hspace{0.02\linewidth}
    \begin{minipage}{0.48\linewidth}
        \centering
        \includegraphics[width=\linewidth]{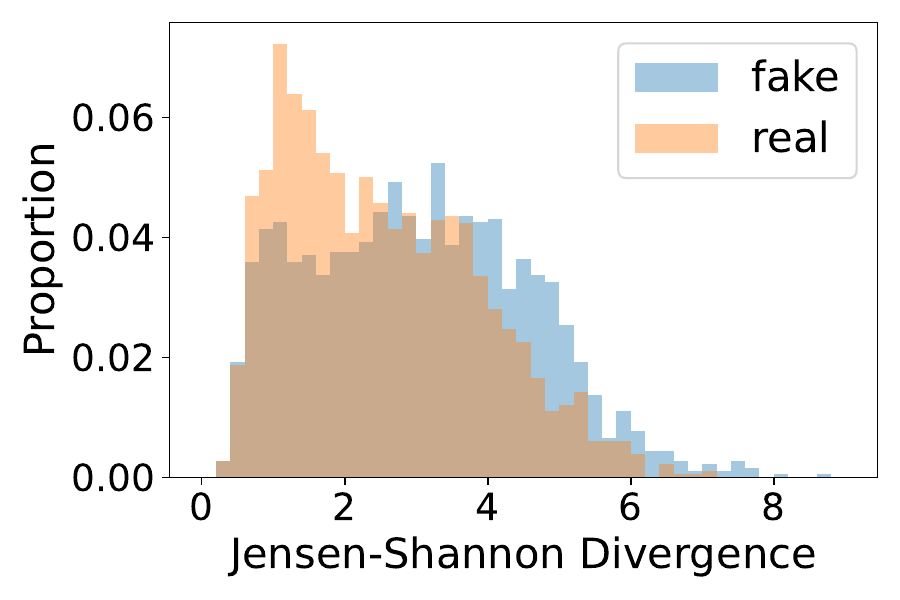}
        \vspace{-0.6cm}
        \caption{JS divergence between textual and visual materials.}
        \label{fig:tv_jsd_fakesv}
    \end{minipage}
\end{figure}

\section{Empirical Analysis} \label{sec: empirical_analysis}
We exhibit the manifestation differences between real and fake news videos in different phases of news video creation by conducting empirical analysis on real-world datasets, including the publicly available Chinese dataset FakeSV~\cite{qi2022fakesv} and the newly constructed English dataset FakeTT.
We identify the discrepancies between real and fake news production processes and provide plausible explanations for these phenomena, highlighting the nuances in the creative process to evaluate the news video veracity. Considering that consistent results were observed across both datasets, we only present findings from FakeSV here due to space limitations and attach results on FakeTT in the supplementary material.
\subsection{Phase \uppercase\expandafter{\romannumeral1} : Material Selection}
\begin{obs}
When selecting audio materials, fake news tends to opt for more emotionally charged audio.
\end{obs}

Considering background music (BGM) is a predominant option for short video news creators and the nature of BGM it serves primarily to evoke emotional responses, our analysis of audio selection behaviors focuses on the emotional aspect. We leverage the pre-trained wav2vec~\cite{speechbrain} that has been fine-tuned for audio emotion classification.
Depicted in Figure~\ref{fig:AudioEmo_fakesv}, we can see that fake news videos exhibit an inclination towards using emotionally charged audio.
Given that prior work~\cite{dobele2007pass} has indicated emotionality significantly boosts content sharing, we attribute this bias in audio selection to creators' intentions to maximize viewer engagement.

\begin{obs}
When selecting visual materials, fake news often employs clips that exhibit a relatively lower semantic consistency with the accompanying text.
\end{obs}
We analyze creators' visual selection behaviors from the perspective of consistency between selected video materials and accompanying text. 
Specifically, we leverage the pre-trained text-image representation model CLIP~\cite{radford2021clip} to extract textual and visual features. By normalizing these features and calculating the Jensen-Shannon (JS) Divergence between the attached text and each frame's visual content, we derive an average JS Divergence score across multiple frames as an indicator of text-visual consistency for the entire video. A lower indicator value signifies a higher semantic consistency between the video's textual and visual content. ~\autoref{fig:tv_jsd_fakesv} illustrates the distinct distributions of JS Divergence between textual and visual materials for fake and real news. The discrepancies have been statistically confirmed through the Kolmogorov-Smirnov (KS) test, with a p-value of less than 0.05. 
We find that fake news tends to utilize visual clips with noticeably lower semantic consistency with the accompanying text.
We attribute the observed biases in video material selection to the nature that fabricated news inherently lacks access to a rich array of related video materials.

\begin{figure}[t]
    \centering
    \begin{minipage}{0.48\linewidth}
        \centering
        \includegraphics[width=\linewidth]{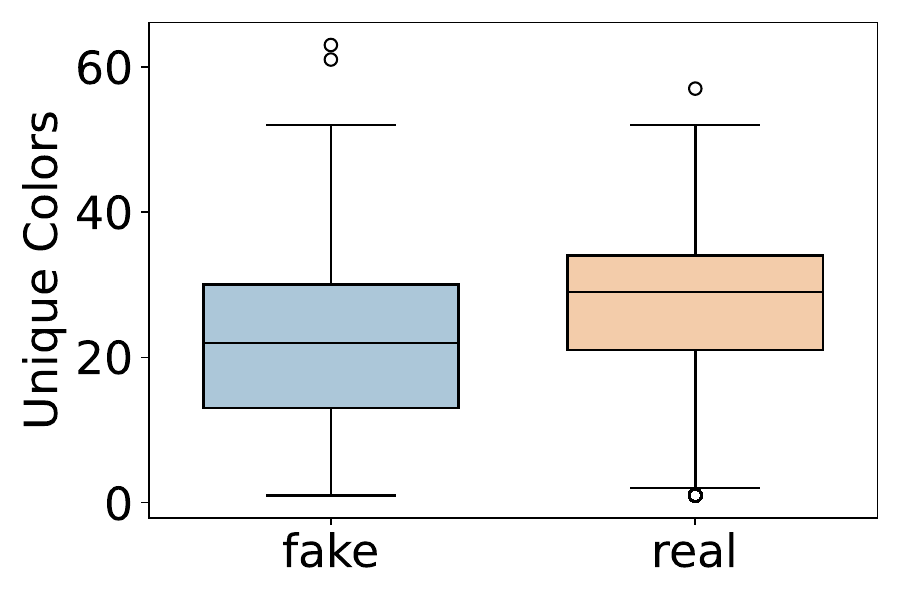}
        \caption{Color richness of on-screen text.}
        \label{fig:ocr_colors_fakesv}
    \end{minipage}%
    \hspace{0.02\linewidth}
    \begin{minipage}{0.48\linewidth}
        \centering
        \includegraphics[width=\linewidth]{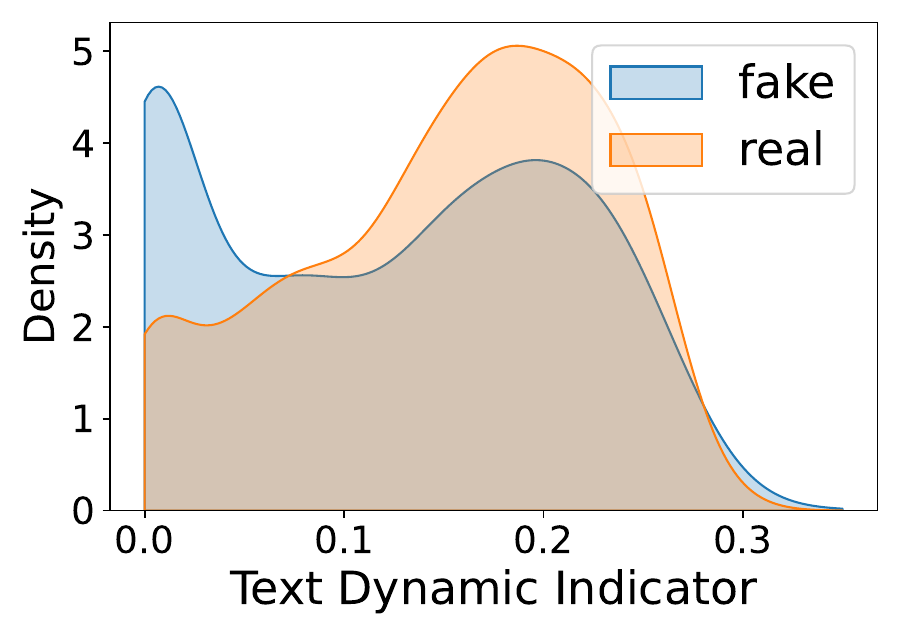}
        \caption{On-screen Text Dynamics.}
        \label{fig:ocr_dynamic_fakesv}
    \end{minipage}
\end{figure}

\begin{figure*}[t]  
    \centering    
    \includegraphics[width=0.9\linewidth]{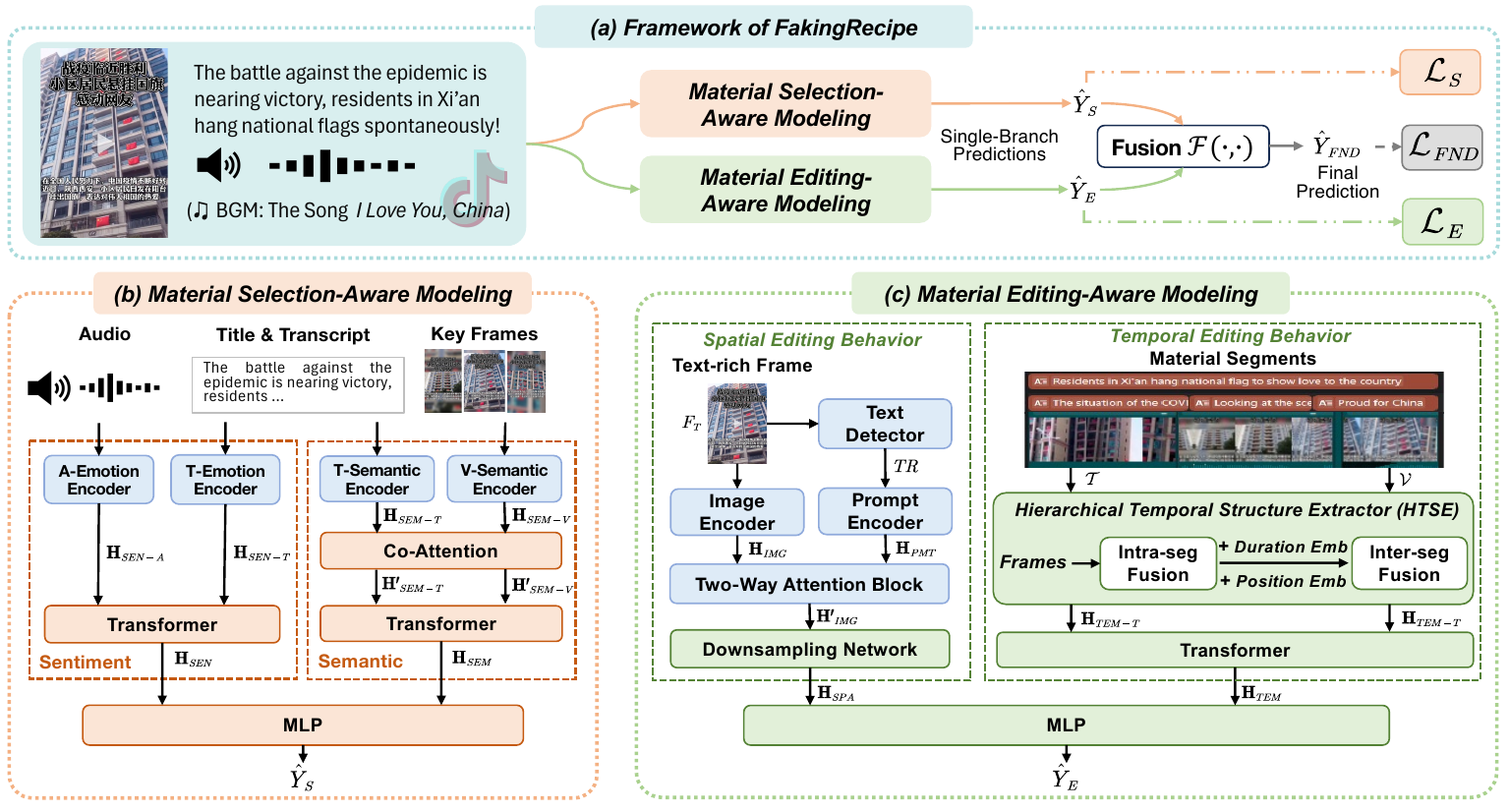}
    \caption{Overview of the proposed FakingRecipe model. (a) Overall framework:  The news video is processed through dual perspectives, with a late fusion strategy employed to integrate clues for final prediction.  (b) Material Selection-Aware Modeling~(MSAM) module: Extracts clues from both sentimental and semantic aspects.  (c) Material Editing-Aware Modeling~(MEAM) module: Extracts clues based on spatial and temporal aspects. $\mathcal{F}(\cdot,\cdot)$ denotes the fusion function. The parameters in the modules in blue are frozen and others are trainable. The overall model is trained under the supervision of the loss functions $\mathcal{L}_{FND},  \mathcal{L}_S, \text{and}  \mathcal{L}_E$. The text in this case is translated into English.}
    \label{fig:model} 
\end{figure*}

\subsection{Phase \uppercase\expandafter{\romannumeral2} : Material Editing}
We investigate two fundamental editing operations in video creation: spatial editing behaviors and temporal editing behaviors. 
\begin{obs}
When spatially imposing text, fake news tends to display relatively plain textual visuals.
\end{obs}
Spatial editing behaviors involve overlaying additional layers on top of the original visual materials. This can include adding animated stickers, text, and other elements. Among them, text imposition is a widely used operation in short news videos (85\% in the FakeSV dataset), with variations reflected in decisions regarding the text's placement, color, typeface, and font. Here we quantified the color characteristics of the text visual areas in real and fake news videos respectively to explore the differences in color choice behaviors during text imposition. ~\autoref{fig:ocr_colors_fakesv} illustrates that real news videos tend to use a richer color palette for text presentation. We attribute this preference to that real news creators often follow conventional editorial norms and invest more effort to improve the presentation quality. Conversely, fake news creators often employ a monochromatic color scheme when imposing text, likely due to a lack of expertise in news production, leaving them unaware of the potential impact these details can have on viewers.

\begin{obs}
When temporally splicing materials, fake news tends to adopt a relatively simple arrangement.
\end{obs}
Temporal editing behaviors, on the other hand, refer to the reorganization and splicing of multiple material segments. The duration and positioning of different segments can subtly influence viewers' perceptions of the news video. Here we examine the temporal editing behaviors related to text exposure, analyzing differences in the temporal arrangement of text segments between real and fake news. Specifically, we developed an indicator, $I_D$, to measure the dynamism of text presentation. By calculating the mean ($\mu$) and standard deviation ($\sigma$) of exposure durations ($d_1$, $d_2$, ...) for different text phrases within a video, $I_D$ is defined as ${\sigma(1-\mu)}$, based on the principle that shorter exposure times and greater variance among text exposure durations indicate stronger text temporal editing dynamism. \autoref{fig:ocr_dynamic_fakesv} shows the fitted sample density distribution of the on-screen text dynamic scores in the FakeSV dataset, revealing significant differences between the temporal editing behaviors of real and fake news, with real news exhibiting more dynamic text presentations. We ascribe this tendency to two factors: First, the disparity in video creation capabilities, wherein most creators of authentic news, endowed with professional media training, possess a nuanced understanding of effectively integrating text with visual elements. Second, the constraints posed by the availability of materials. Fabricated news, inherently characterized by its scant and biased content, often lacks the robust information necessary for dynamic presentations. This deficiency compels creators to resort to the static placement of limited information in specific areas of the screen.

\section{Method}

\subsection{Overview}
Drawing on the insights from our empirical analysis, we present FakingRecipe (\autoref{fig:model}), a creative process-aware fake news video detection model. The model observes the given news video from the two pivotal phases of the creative process to unearth veracity indicating clues. 
Treating the feature from two phases as independent viewpoints, FakingRecipe is structured with dual branches operating separately and employs a late fusion strategy to integrate predictions from these independent perspectives.

\subsection{Material Selection-Aware Modeling~(MSAM)}
Based on prior analysis, we examine the creators' material selection behavior from two aspects (\textit{i.e.}, sentiment and semantic). The dominant role of different modalities varies in conveying information: Audio primarily expresses emotion, text renders emotional tones while conveying semantic information, and visuals generally complement the text to communicate semantic content. Therefore, we strategically select combinations of modalities for multifaceted feature extraction, subsequently fusing multimodal features from multiple viewpoints.

Specifically, for the sentimental aspect, we consider audio and textual content as the primary sources. We utilize fine-tuned versions of HuBERT~\cite{hsu2021hubert} and XLM-RoBERTA~\cite{conneau2019xlmroberta} as encoders to extract audio sentimental features $\textbf{H}_{SEN-A}$ and textual sentimental features $\textbf{H}_{SEN-T}$, respectively. These sentimental features from different modalities are then concatenated and fed into a standard Transformer layer~\cite{vaswani2017attention}. By leveraging self-attention, the transformer layer fuses multimodal sentimental features to produce a unified sentimental feature representation $\textbf{H}_{SEN}$.

In the semantic aspect, visual and textual contents take precedence, while the audio mainly serves as background music, playing a minimal role.  
Keyframes are extracted from videos, serving as the basis for visual analysis.  Utilizing CLIP~\cite{radford2021clip}, we encode text and keyframes to token/frame-level text semantic features $\textbf{H}_{SEM-T}$ and visual semantic features $\textbf{H}_{SEM-V}$.  Interaction between text and visual content is facilitated through a co-attention transformer~\cite{lu2019coattn}, resulting in visually enhanced textual features  $\textbf{H'}_{SEM-T}$ and textually enhanced visual features $\textbf{H'}_{SEM-V}$.  These features are then averaged, concatenated, and input into a transformer layer mirroring the structure used in the sentimental analysis. This process integrates semantic features from various modalities into a singular semantic feature representation $\textbf{H}_{SEM}$.

The sentimental feature $\textbf{H}_{SEN}$ and semantic feature $\textbf{H}_{SEN}$ are then concatenated and fed into a two-layer MLP to derive the fake news predicted score $\hat{Y}_S$ from the material selection analysis perspective:
\begin{equation}
    \hat{Y}_S=\text{MLP}([\textbf{H}_{SEN};\textbf{H}_{SEM}]).
\end{equation}
\subsection{Material Editing-Aware Modeling~(MEAM)}
In mining detecting clues from the perspective of creator editing behaviors, we focus on spatial and temporal aspects, identified as critical in our empirical analysis.

\textbf{Spatially}, we examine the prevalent practice of imposing text. Given a video $V$, we select a representative text-rich frame $F_T$, identified based on the size of the text presence area, as our starting point. We first employ an OCR spotting model, CRAFT~\cite{baek2019CRAFT}, to delineate text regions $TR=\{\text{box}_1,\text{box}_2,...\}$ within $F_T$. These regions are subsequently transformed into prompt embeddings $\textbf{H}_{PMT}$ employing a methodology inspired by the prompt encoder in Segment Anything Model (SAM)~\cite{kirillov2023SAM}.  In parallel, $F_T$ undergoes processing by a pre-trained Vision Transformer (ViT)~\cite{dosovitskiy2020image} to produce initial encodings $\textbf{H}_{IMG}$. 
Both $\textbf{H}_{IMG}$ and $\textbf{H}_{PMT}$ are then fed into a Two-Way Attention block to refine the initial visual encoding $\textbf{H}_{IMG}$, ensuring it focuses more accurately on text regions within the frame. 
Suppose the attention mechanism is described as $\text{Att}( \mathbf{Q},  \mathbf{K},  \mathbf{V}) = \text{softmax} ( \mathbf{Q'}\cdot\mathbf{K'}/\sqrt{d})  
 \mathbf{V'}$, where $ \mathbf{Q'} = \mathbf{W}_Q  \mathbf{Q}$, $ \mathbf{K'} =  \mathbf{W}_K  \mathbf{K}$, $ \mathbf{V'} =  \mathbf{W}_V  \mathbf{V}$, and $d$ is the dimensionality. We format self-attention as $\text{SA}(\mathbf{X})=\text{Att}( \mathbf{X}, \mathbf{X}, \mathbf{X})$ and cross-attention as $\text{CA}( \mathbf{X},  \mathbf{Y})=\text{Att}( \mathbf{X}, \mathbf{Y}, \mathbf{Y}) $. The Two-Way Attention block functions in two directions as follows:
\begin{align}
\vspace{-4mm}
& \mathbf{H\mbox{-}H}_{PMT} = \text{LN}(\mathbf{H}_{PMT}+\text{SA}(\mathbf{H}_{PMT})), && \\
& \mathbf{H'\mbox{-}H}_{PMT} = \text{LN}(\mathbf{H\mbox{-}H}_{PMT} + \text{CA}(\mathbf{H\mbox{-}H}_{PMT}, \mathbf{H}_{IMG})), && \\
& \mathbf{H''\mbox{-}H}_{PMT} = \text{LN}(\mathbf{H'\mbox{-}H}_{PMT} + \text{MLP}(\mathbf{H'\mbox{-}H}_{PMT})), && \\
& \mathbf{H'}_{IMG} = \text{LN}(\mathbf{H}_{IMG} + \text{CA}(\mathbf{H}_{IMG}, \mathbf{H''\mbox{-}H}_{PMT})),
\vspace{-4mm}
\end{align}
\noindent where LN represents layer normalization. This block uses an embedding dimension of 256, and all attention layers use 8 heads. Following SAM, we adopt two layers of such block to obtain the updated image feature $\mathbf{H'}_{IMG}$ that focuses on text regions.  \\
Following attention processing, the updated $\textbf{H'}_{IMG}$ undergoes downsampling via two convolutional layers and then flattened to derive the spatial pattern feature $\textbf{H}_{SPA}$:
\begin{equation} 
\mathbf{H}_{SPA} = \text{GeLU}(\text{Conv}(\text{GeLU}(\text{LN}(\text{Conv}(\mathbf{H'}_{IMG}))))).
\end{equation}

\textbf{Temporally}, we examine the splicing practice of text segment and video segment. The audio track is omitted due to the observation that most audio tracks consist of continuous background music. For a Video $V$, preprocessing extracts a sequence of text content $\mathcal{T}=\{(t_1,d_1),(t_2,d_2),...(t_n,d_n)\}$ and a sequence of visual content $\mathcal{V}=\{(v_1,d_1),(v_2,d_2),...(v_m,d_m)\}$, with $n$ and $m$ indicating the counts of text and video segments respectively. $t_i$ represents the $i$-th textual segment, $v_i$ denotes the middle frame of the $i$-th video clip, and $d_i=[\text{FrameIdx}^{begin}_i, \text{FrameIdx}^{end}_i]$ marks the time interval of the $i$-th segment's appearance. The input also incorporates frame rate  $(\text{fps})$ and the total frame count $(\text{vframes})$ of the video to contextualize duration. Each modality’s temporal structure is initially modeled separately, followed by an interaction phase to derive overall temporal editing features. Specifically, we design a Hierarchical Temporal Structure Extractor (HTSE) for extracting temporal structure features applicable to both modalities. HTSE first performs intra-segment fusion for content occurring in the same time span to derive segment content features $\textbf{Seg}_i$.  For text, $\textbf{Seg}_i^{T}$ is obtained by concatenating multiple segments and encoding them collectively, while for visuals, it applies the self-attention ($SA$) mechanism for integration:
\begin{equation}
    \textbf{Seg}_i^{V}=\text{MEAN}(\text{SA}([v_1,v_2,...,v_k])), 
\end{equation}
where k is the frame count within segment $i$, and $\text{MEAN}(\cdot)$ denotes the mean pooling. To model the subtle influences of the duration and temporal position of different segments, we introduce each segment's temporal position and exposure duration information. Positional encoding ($\textbf{PE}$) is generated using sine and cosine functions to reflect each segment’s temporal position, akin to that leveraged by Transformer~\cite{vaswani2017attention}:
\begin{equation}
    \text{PE}_i^{(ei)}=\left\{\begin{array}{lc}
    \sin \left(w_k i\right), \quad \text { if } ei=2 k \\
    \cos \left(w_k i\right), \quad \text { if } ei=2 k+1 ,
    \end{array}\right.
\end{equation}
where $w_k=1/(10000^{2 k / \text{dim}_{PE}})$ represents the frequency of the sinusoid for each dimension and $\textbf{PE}_i$ is the $i$-th segment's positional embedding. For duration encoding (\textbf{DE}), we employ an equi-frequency binning approach, determining duration groups through empirical analysis and assigning a learnable embedding to each group. Considering the audience's perception of exposure time in reality, both absolute and relative durations are evaluated:
\begin{align}
    \text{Dura}^{abs}_i &=(\text{FrameIdx}^{begin}_i- \text{FrameIdx}^{end}_i)/\text{fps} , \notag\\
    \text{Dura}^{rel}_i &=(\text{FrameIdx}^{begin}_i- \text{FrameIdx}^{end}_i)/\text{vframes} . \notag
\end{align}
Here $\text{Dura}^{abs}_i$ and $\text{Dura}^{rel}_i$  denote the absolute (in seconds) and relative (the proportion of the total duration) durations of segment $i$, respectively. The $i$-th segment's duration embedding is:
\begin{equation}
    \textbf{DE}_i=[\text{Emb}(\text{Group}(\text{Dura}^{abs}_i));\text{Emb}(\text{Group}(\text{Dura}^{rel}_i))],
\end{equation}
where $\text{Group}(\cdot)$ maps a duration to its designated group and $\text{Emb}(\cdot)$ retrieves the corresponding embedding for that group. 

Integrating positional and duration encodings, the segment features are updated to $\textbf{SEG}_i$, serving as the input for inter-segment fusion, which captures the relationships between different segments using a similar self-attention mechanism, generating temporal pattern features for each modality:
\begin{align}
    \textbf{SEG}_i &=\textbf{Seg}_i+\textbf{PE}_i+\textbf{DE}_i , \\
    \textbf{H}_{TEM-M} &=\text{MEAN}(\text{SA}([\textbf{SEG}_1^M,\textbf{SEG}_2^M,...])).
\end{align}
Utilizing HTSE,  we derive temporal editing features for both text $(\textbf{H}_{TEM-T})$ and visual $(\textbf{H}_{TEM-V})$ modalities, and they are subsequently processed by a standard Transformer layer to produce the consolidated temporal editing feature $\textbf{H}_{TEM}$. The spatial editing feature $\textbf{H}_{SPA}$ and the temporal editing feature $\textbf{H}_{TEM}$ are then concatenated and fed into a two-layer MLP to compute the fake news predicted score $\hat{Y}_E$ from the material editing analysis perspective:
\begin{equation}
\hat{Y}_E=\text{MLP}([\textbf{H}_{SPA};\textbf{H}_{TEM}] .
\end{equation}
\subsection{Predication and Optimization} 
\subsubsection{Prediction}
Building on the predicted scores $\hat{Y}_S$ from material selection modeling and $\hat{Y}_E$ from material editing modeling, we adopt a late fusion strategy to get the final score $\hat{Y}_{FND}$:
\begin{equation}
    \hat{Y}_{FND}=\mathcal{F}(\hat{Y}_S,\hat{Y}_E)=\hat{Y}_S * \tanh(\hat{Y}_E),
\end{equation}
where $\mathcal{F}(\cdot, \cdot)$ is the fusion function. Inspired by previous works~\cite{wang2021clicks, chen2023causal}, we adopt the $\tanh(\cdot)$ function, which introduces non-linearity to enhance the fusion strategy's representational capacity.

\subsubsection{Optimization} Following previous works~\cite{qi2022fakesv, shang2021tiktec, choi2021cikm}, we utilize cross-entropy loss to optimize our model:
\begin{equation}
    \mathcal{L}_{FND} = \text{Cross-Entropy}(\hat{Y}_{FND}, Y),
\end{equation}
where $Y$ is the ground-truth label for each short video news.

To further supervise the material selection and material editing modeling, the final loss $\mathcal{L}$ incorporates the loss for $\hat{Y}_S$ and $\hat{Y}_E$:
\begin{equation}
  \mathcal{L}=\mathcal{L}_{FND}+\alpha \mathcal{L}_S +\beta \mathcal{L}_E, 
\end{equation}
where $\alpha$ and $\beta$ are hyperparameters that balance the impacts on the back-propagation of the three branches. $\mathcal{L}_S$ and $\mathcal{L}_E$ denote the $\text{Cross-Entropy}(\hat{Y}_S, Y)$ and $\text{Cross-Entropy}(\hat{Y}_E, Y)$, respectively.

\begin{table}[t]
\centering
\caption{Statistics of two datasets for evaluation.}
\vspace{-0.2cm}
\label{dataset_statistic}
\resizebox{1\columnwidth}{!}{
\begin{tabular}{cccrrr} 
\toprule
Dataset & Time Range & Avg Duration (s) & \#Fake & \#Real & \#All  \\ 
\midrule
FakeSV  & 2017/10-2022/02 & 39.88       & 1,810   & 1,814   & 3,624   \\
FakeTT  & 2019/05-2024/03 & 47.69       & 1,172   & 819    & 1,991   \\
\bottomrule
\end{tabular}
}
\end{table}

\begin{table*}[t]
\centering
\caption{Performance comparison between FakingRecipe and baselines on the FakeSV and FakeTT datasets. The best performance in each column is bolded and the relative improvement of FakingRecipe over the best baseline is in the brackets.}
\vspace{-0.2cm}
\label{comparison}
\resizebox{0.85\textwidth}{!}{
\begin{tabular}{cccccccccc} 
\toprule
\multirow{2}{*}{Dataset} & \multirow{2}{*}{Method} & \multirow{2}{*}{Accuracy} & \multirow{2}{*}{Macro F1} & \multicolumn{3}{c}{Fake}                    & \multicolumn{3}{c}{Real}                     \\ 
\cline{5-10}
                         &                         &                           &                           & Precision      & Recall         & F1       & Precision      & Recall         & F1        \\ 
\midrule
\multirow{8}{*}{FakeSV}  & GPT-4                   & 67.43                     & 67.34                     & 83.71          & 53.99          & 65.64          & 57.81          & \textbf{85.71} & 69.05           \\
                         & GPT-4V                  & 69.15                     & 69.14                     & 82.35          & 58.78          & 68.60          & 60.00          & 83.08          & 69.68           \\ 
\cdashline{2-10}
                         & HCFC-Hou~               & 74.91                     & 73.61                     & 73.46          & 86.51          & 79.46          & 77.72          & 60.08          & 67.77           \\
                         & HCFC-Medina             & 76.38                     & 75.83                     & 77.50          & 81.58          & 79.49          & 74.77          & 69.75          & 72.17           \\ 
\cdashline{2-10}
                         & FANVM~                  & 79.52                     & 78.81                     & 78.64          & 87.17          & 82.68          & 80.98          & 69.75          & 74.94           \\
                         & TikTec~                 & 73.43                     & 73.26                     & 78.37          & 72.70          & 75.43          & 68.08          & 74.37          & 71.08           \\
                         & SVFEND~                 & 80.88                     & 80.54                     & \textbf{85.82} & 77.63          & 81.52          & 74.53          & 83.61          & 78.81           \\ 
\cdashline{2-10}
                         & \textbf{FakingRecipe~(Ours)}  & $\textbf{85.35}_{(+5.53\%)}$           & $\textbf{84.83}_{(+5.33\%)}$          & 83.33          & \textbf{92.11} & \textbf{87.50} & \textbf{88.35} & 76.47          & \textbf{81.98}  \\ 
\midrule
\multirow{8}{*}{FakeTT}  & GPT-4                   & 61.45                     & 60.66                     & 43.36          & 75.61          & 55.11          & 83.19          & 55.00          & 66.22           \\
                         & GPT-4V                  & 58.69                     & 58.69                     & 44.52          & \textbf{88.46} & 59.23          & 88.00          & 43.42          & 58.15           \\ 
\cdashline{2-10}
                         & HCFC-Hou~               & 73.24                     & 72.00                     & 56.93          & 78.79          & 66.10          & 87.04          & 70.50          & 77.90           \\
                         & HCFC-Medina             & 62.54                     & 62.23                     & 46.24          & 80.81          & 58.82          & 84.92          & 53.50          & 65.64           \\ 
\cdashline{2-10}
                         & FANVM~                  & 71.57                     & 70.21                     & 55.15          & 75.76          & 63.83          & 85.28          & 69.50          & 76.58           \\
                         & TikTec~                 & 66.22                     & 65.08                     & 49.32          & 72.73          & 58.78          & 82.35          & 63.00          & 71.39           \\
                         & SVFEND~                 & 77.14                     & 75.63                     & 62.33          & 78.79          & 69.57          & 87.91          & 76.33          & 81.69           \\ 
\cdashline{2-10}
                         & \textbf{FakingRecipe~(Ours)}  & $\textbf{79.15}_{(+2.61\%)}$            & $\textbf{77.74}_{(+2.79\%)}$           & \textbf{64.75} & 81.82          & \textbf{72.18} & \textbf{89.74} & \textbf{77.83} & \textbf{83.30}  \\
\bottomrule
\end{tabular}
}
\end{table*}

\section{Experiments}
In this section, we conduct extensive experiments on two real-world datasets to verify the effectiveness of FakingRecipe by comparing it with seven representative baselines and the FakingRecipe variants.

\subsection{Datasets}
To validate the generalizability of the proposed FakingRecipe, we conduct experiments on two datasets of different languages:

\textbf{FakeSV}\footnote{\url{https://github.com/ICTMCG/FakeSV}}: The largest publicly available Chinese dataset for fake news detection on short video platforms, featuring samples from \textit{Douyin} and \textit{Kuaishou}, two popular Chinese short video platforms. Each sample in FakeSV contains the video itself, its title, comments, metadata, and publisher profiles. We do not use the last three values to focus on understanding the content itself.

\textbf{FakeTT}: Our newly constructed English dataset for a comprehensive evaluation in English-speaking contexts
\footnote{\citet{shang2021tiktec} did collect an English TikTok dataset but did not release it. We did not receive any reply to our email for the dataset inquiry.}.
Curated from TikTok, this dataset follows a similar collection process to~\cite{qi2022fakesv}, focusing on videos related to events reported by the fact-checking website Snopes\footnote{\url{https://www.snopes.com/}}. Each video was rigorously annotated for authenticity by at least two independent annotators, resulting in a collection of 1,172 fake news videos and 819 real news ones from May 2019 to March 2024, with video, audio, and text description (title) available. See more details in the supplementary material.

Table~\ref{dataset_statistic} shows the statistics of the two datasets. To simulate real-world scenarios, we adopt a temporal split strategy for our experiments, dividing the data chronologically into training, validation, and testing sets with ratios of 70\%, 15\%, and 15\%, respectively. 
Such a data split reflects the potential of applying compared methods in reality.

\subsection{Experimental Setup}
We compare the proposed FakingRecipe with a range of state-of-the-art baselines, including handcrafted features-based baselines, neural networks-based baselines, and (multimodal) large language model  ((M)LLMs) baselines:

\textbf{Handcraft Feature-based Baselines:}
(1)~\textbf{HCFC-Hou~\cite{hou2019icmi}} 
utilizes linguistic features from speech, acoustic emotion features, and user engagement statistics with a linear kernel SVM for classification.
(2)~\textbf{HCFC-Medina~\cite{aclworkshop2020}} extracts TF-IDF vectors from video titles and the first hundred comments, applying SVM for detection.

\textbf{Neural Network-based Baselines:}
(1)~\textbf{FANVM~\cite{choi2021cikm}} 
harnesses visual features from keyframes and textual features from titles and comments, using an adversarial network to extract topic-agnostic multimodal features for classification.
(2)~\textbf{TikTec~\cite{shang2021tiktec}}
employs speech text-guided visual object features and MFCC-guided speech textual features, using a co-attention mechanism for fusion and classification.
(3)~\textbf{SVFEND~\cite{qi2022fakesv}} 
leverages cross-modal transformers to boost interaction between modalities and integrates content with social context features via a self-attention mechanism.

\textbf{(M)LLM Baselines:}
(1)~\textbf{GPT-4~\cite{gpt4}} is one of the most powerful LLMs currently and is used to make predictions based on video news titles and extracted on-screen text. We use a zero-shot prompt template inspired by \citet{hu2024bad}.
(2)~\textbf{GPT-4V~\cite{GPT-4V}} 
is the variant of GPT-4 that supports visual inputs. We include the video’s thumbnail in the inputs to explore the capabilities of (M)LLMs in this task.

Given that we focus on content-only detection at the early news spreading stage, all baselines are adapted to rely solely on content.

For the implementation details and metrics, please refer to the supplementary material.

\subsection{Overall Performance}
Table~\ref{comparison} presents the performance of FakingRecipe and the compared baselines. The results reveal several key observations:

First, the zero-shot (M)LLM-based methods, namely GPT-4(V), underperform the methods specifically tailored for fake news video detection, which indicates the complexity of the task and the necessity of specialized models for this task currently. Notably, GPT-4(V) exhibits biases in authenticity judgments, tending to classify videos as real in the FakeSV dataset and as fake in the FakeTT dataset, possibly due to different knowledge accumulation in the tested large models.

Second, neural network-based baselines generally outperform handcraft feature-based baselines, demonstrating the superiority of automated neural network models in handling complex fake news detection tasks. However, in some instances, the handcraft feature-based baselines surpass certain neural network models, particularly TikTec. This suggests that integrating human-guided knowledge into the models may bring additional advantages in specific cases.

Finally, FakingRecipe outperforms all competing methods in Accuracy and macro F1 on both datasets, validating its effectiveness in detecting fake news videos. Notably, the improvements are more pronounced on the FakeSV dataset (5.53\% increase in accuracy and 5.33\% in macro F1) compared to that on FakeTT (2.61\% in accuracy and 2.79\% in macro F1), possibly reflecting the moderate pattern differences of the creative process in different cultural background.

\begin{table}[t]
\small
\caption{Ablation study of multiple model components.}
\vspace{-0.2cm}
\label{tab:ablation}
\begin{tabular}{cccc|cccc}
\hline
\multicolumn{4}{c|}{Module}                          & \multicolumn{4}{c}{Dataset}                                     \\ 
\hline
\multicolumn{2}{c}{MSAM} & \multicolumn{2}{c|}{MEAM} & \multicolumn{2}{c|}{FakeSV}        & \multicolumn{2}{c}{FakeTT} \\
SEN         & SEM        & SPA         & TEM         & Acc   & \multicolumn{1}{c|}{F1}    & Acc          & F1          \\ 
\hline
\usym{2713}           & \usym{2713}          & \usym{2713}           & \usym{2713}           & \textbf{85.35} & \multicolumn{1}{c|}{\textbf{84.83}} & \textbf{79.15}        & \textbf{77.74}       \\
\cdashline{1-8}
\usym{2713}           & \usym{2713}          &             &             & 83.94 & \multicolumn{1}{c|}{83.56} & 77.92            &  76.61           \\
            &            & \usym{2713}           & \usym{2713}           & 82.47 & \multicolumn{1}{c|}{81.96} & 71.24        & 69.81       \\
\cdashline{1-8}
            & \usym{2713}          & \usym{2713}           & \usym{2713}           & 83.58 & \multicolumn{1}{c|}{83.10}  & 76.92        & 75.95       \\
\usym{2713}           &            & \usym{2713}           & \usym{2713}           & 84.31 & \multicolumn{1}{c|}{83.92} & 74.91        & 73.71       \\
\usym{2713}           & \usym{2713}          &             & \usym{2713}           & 84.87 & \multicolumn{1}{c|}{84.42} &   78.76           &   77.39          \\
\usym{2713}           & \usym{2713}          & \usym{2713}           &             & 84.14 & \multicolumn{1}{c|}{83.81} & 78.59        & 77.53            \\ 
\hline
\end{tabular}
\end{table}

\subsection{Ablation Study}

To rigorously evaluate the individual contributions of each component within FakingRecipe, we conduct extensive ablation studies, the results of which are detailed in \autoref{tab:ablation}. We first focus on the performances of the two core modules: Material Selection-Aware Modeling (MSAM) and Material Editing-Aware Modeling (MEAM). It is observed that MSAM generally outperforms MEAM, with a more notable performance gap observed on the FakeTT dataset compared to FakeSV. While MEAM showed relatively lower performance on its own, it provides crucial complementary insights that significantly enhance the overall effectiveness of the combined model beyond what is achieved by MSAM alone.

Further exploration into each specific aspect within these modules: sentimental and semantic for MSAM, and spatial and temporal for MEAM. By systematically removing each aspect and comparing the altered model's performance to the original, the results confirm that each component plays a vital role in the model's overall effectiveness. Among these aspects, while the spatial component shows the smallest improvement in performance, the sentimental aspect is most impactful for FakeSV, and the semantic aspect is particularly effective for FakeTT. This detailed analysis not only demonstrates the essential contribution of each aspect but also underscores the synergy that their integration brings to the effectiveness of FakingRecipe in detecting fake news videos.

\subsection{Further Analysis}

The performance improvements in FakingRecipe are attributed to the enhancements by the MSAM and MEAM modules. Specifically, MSAM facilitates multimodal content understanding and MEAM introduces a novel perspective on mining content utilization. In this section, we conduct a deeper investigation into these two modules and present two additional findings:

\begin{figure}[t]
    \centering
    \vspace{-0.3cm}
    \subfigure[FakeSV]{
    \includegraphics[width=0.22\textwidth]{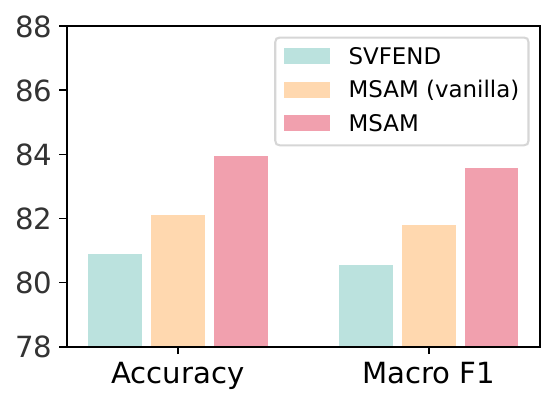}
    }
    \subfigure[FakeTT]{
    \includegraphics[width=0.22\textwidth]{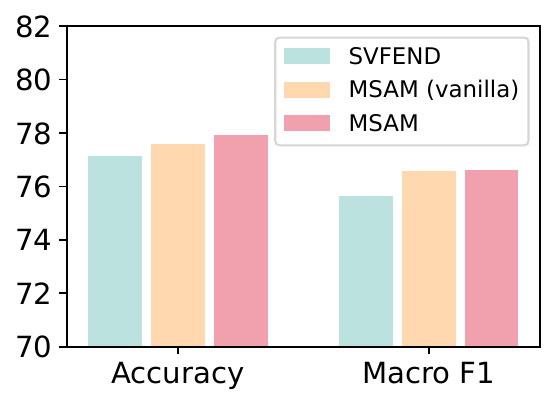}
    }
    \vspace{-0.2cm}
    \caption{Performance comparison of the proposed MSAM module and the best baseline SVFEND on the FakeSV and FakeTT datasets.}
    \label{fig: efficacy_msam}
\end{figure}

\textbf{The synergy of creative process-aware encoding and fusion strategy deepens understanding of video materials, leading to better detection performance.}
We implement a simplified version of MSAM, termed MSAM (vanilla), which directly concatenates features from multiple encoders for classification. As depicted in \autoref{fig: efficacy_msam}, MSAM (vanilla) performs better than SVFEND which employs universal encoders for multimodal content understanding, confirming the efficacy of our material selection-aware multimodal content encoding strategy. However, its performance still falls behind the full MSAM configuration, emphasizing the crucial role of the advanced fusion structure in performance improvement. This exploration underscores the individual effectiveness of both the encoding and fusion strategy and their synergy within MSAM.
\footnote{The analysis of different fusion strategies is in the supplementary material.}

\textbf{Creative process-aware modeling introduces new effective clues that can even bring improvements to other existing models.}
We assess the generalizability of MEAM, which introduces a novel perspective in modeling multimodal content utilization. We directly integrate MEAM into TikTec and SVFEND using the same late fusion strategy as FakingRecipe. The results on two datasets are shown in \autoref{tab: efficacy_meam}. We see that incorporating MEAM resulted in performance gains on both baselines, with TikTec showing significant improvements, affirming MEAM's capacity to elevate performance under effective fusion.

\begin{table} [t]
\centering
\small
\caption{Performance comparison of different models enhanced by our proposed MEAN module on two datasets.}
\vspace{-0.2cm}
\label{tab: efficacy_meam}
\begin{tabular}{c|cc|cc} 
\hline
\multirow{2}{*}{Method}                & \multicolumn{2}{c|}{FakeSV} & \multicolumn{2}{c}{FakeTT}  \\
                                      & Accuracy & Macro F1         & Accuracy & Macro F1         \\ 
\hline
TikTec                                & 73.43    & 73.26            & 66.22    & 65.08            \\
\multicolumn{1}{r|}{\textit{(+MEAM)}} & 83.95    & 83.52            & 71.57    & 70.61            \\ 
\hdashline
SVFEND                                & 80.88    & 80.54            & 77.14    & 75.63            \\
\multicolumn{1}{r|}{\textit{(+MEAM)}} & 83.03    & 82.37            & 78.76    & 77.15            \\
\hline
\end{tabular}
\vspace{-0.2cm}
\end{table}

\subsection{Case Study}
\begin{figure}[t]  
    \centering    
\includegraphics[width=1\linewidth]{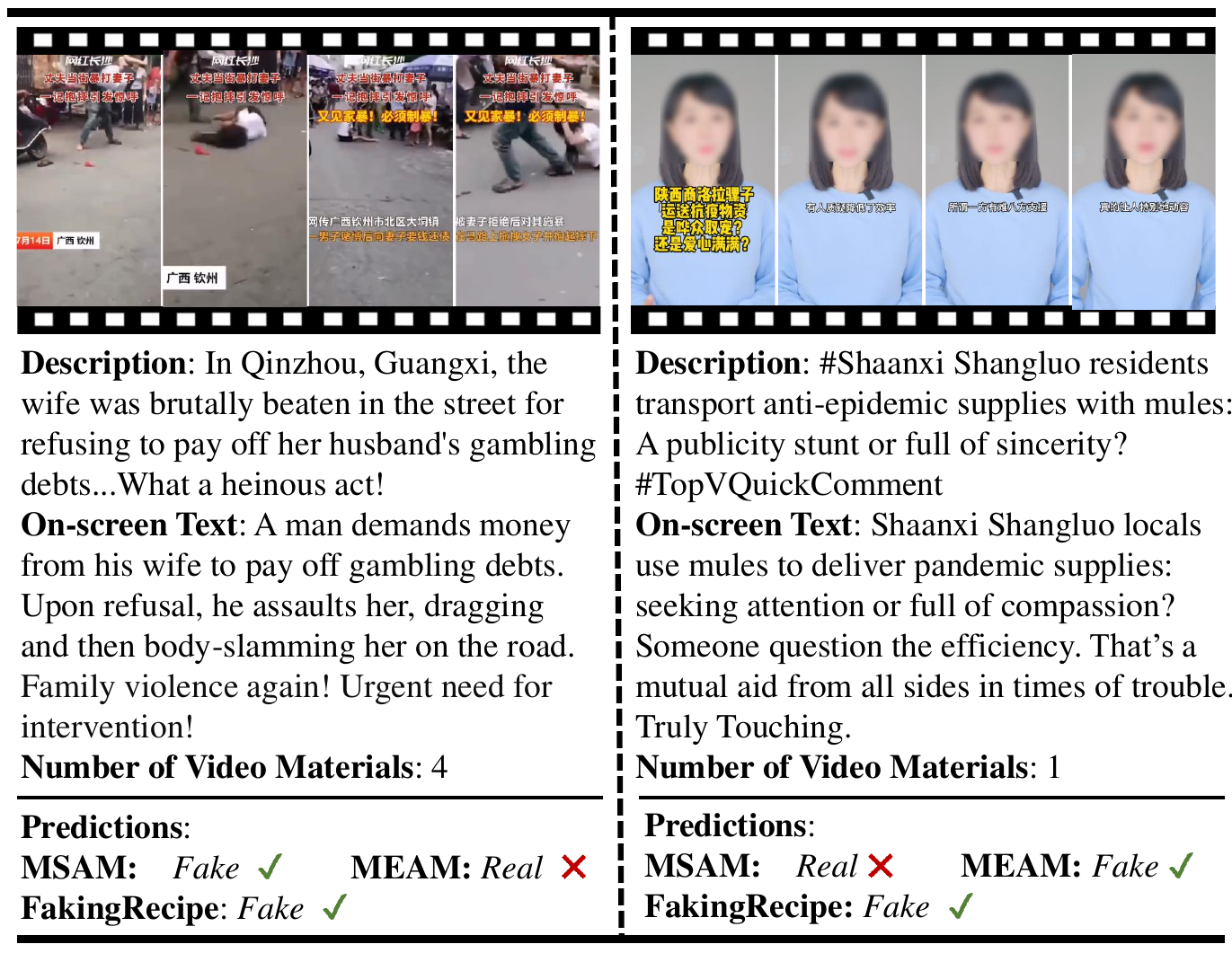}
    \caption{Two fake news cases from FakeSV demonstrating the complementary roles of MSAM and MEAM in FakingRecipe. We translate the texts into English and blur the faces to respect user privacy.}
    \label{fig:case_study} 
    \vspace{-0.5cm}
\end{figure}

We further demonstrate the complementary capabilities of MSAM and MEAM in detecting fake news videos through two real examples from the FakeSV dataset in \autoref{fig:case_study}.
In the left example, a video with high-quality production and visually rich materials is evaluated. Influenced by the video's polished appearance, MEAM classifies it as real. However, MSAM assesses the situation from a different angle, detecting emotionally charged language in the video’s title, such as ``what a heinous act,'' which identifies as a potential indicator of misinformation. This nuanced analysis by MSAM accurately flags the video as fake, showcasing its ability to probe deeper than superficial qualities.
Conversely, the right example presents a video with a neutral expression, which initially leads MSAM to classify it as authentic. Here, MEAM provides critical complementary information. It scrutinizes the video’s sparse visual content and simplistic textual presentation, cues that suggest a lack of authenticity. This focused evaluation by MEAM correctly identifies the video as fake, highlighting its essential role in the overall analysis.
These case studies underscore the complementary nature of MSAM and MEAM in FakingRecipe, enabling a layered and comprehensive assessment of news videos. 

\section{Related Work}
\paratitle{Fake News Video Detection.}
The early work closely related to fake news video detection traces its roots to multimedia forensics research. Forensics-based works follow a basic idea about veracity that misinformation videos are often produced using forgery techniques~\cite{bu2023combating,ganti2022}.
However, with the prevalence of user-friendly editing tools, manipulating visual content has become a common practice across social media platforms, significantly limiting the applicability of this detection approach. Thus, recent investigations have shifted their methodology towards mining detection clues from multimodal content. Handcraft features tailored for fake news video detection~\cite{palod2019vavd,papadopoulou2017,hou2019icmi,aclworkshop2020} like linguistic patterns, acoustic emotion, and user engagement statistics are designed. Further studies incorporate visual expression and leverage deep neural networks~\cite{li2022cnn,jagtap2021misinformation,liu2023covid,qi2022fakesv,choi2021cikm,shang2021tiktec} for falsehood identification. 
Building on the foundation of multimodal content clues within individual samples, some researchers propose to incorporate the neighborhood relationship in an event for fake news video detection, exemplified by the NEED framework~\cite{qi2023two}. Though effective, its dependency on existing data accumulations limits its applicability in real-world scenarios. Instead, our method is suitable for detection at the early stage as it only requires the video content as the input.

\paratitle{Narrative-aware Fake News Detection.}
News reporting has long been seen as involving a form of storytelling~\cite{wahl2019news,bateman2023multimodal,tseng2023fakenarratives}. From this perspective, applying narrative theory, a discipline focusing on how stories are depicted persuasively~\cite{bruner2009actual}, to characterize fake news emerges as an intuitive idea. 
Narrative theory emphasizes analyzing the ``what'' (the content of the story) and the ``how'' (the strategy of storytelling) as its two pivotal aspects~\cite{fludernik2009introduction}, echoing the perspective of the creative process. 
The potential of applying narrative theory for detecting fake news has been demonstrated by studies on news articles~\cite{hamby2024sensational,wang2024framing,karimi2019learning,sheng2021integrating}.
However, within the realm of multimodal news, related research remains limited.
Current studies in multimodal fake news detection~\cite{wang2018eann,qi2021improving,chen2022cross,khattar2019mvae,ying2023bootstrapping,wu2021multimodal} typically concentrate solely on the analysis of presented content, neglecting the broader narrative structures.
\citet{tseng2023fakenarratives} make the first foray into understanding narratives of disinformation in TV news videos. A multimodal discourse analysis scheme is proposed to uncover narrative strategies~\cite{bateman2023multimodal}. However, their focus is to assist manual statistical analysis using web-based tools~\cite{liebl2023designing} and thus is inapplicable to automatic detection. Our study takes the first step to detect fake news on short video platforms from the perspective of the creative process, which can be seen as a practical solution of the narrative theory for this task. 

\section{Conclusion}
We proposed to detect fake news on short video platforms from the perspective of the creative process and designed the creative process-aware detector, FakingRecipe. It observes the given video from material selection and editing perspectives to capture the unique production characteristics of fake news videos. We conducted experiments on the English FakeTT dataset newly constructed by us and the popular Chinese FakeSV dataset and validated the effectiveness of FakingRecipe.

\bibliographystyle{ACM-Reference-Format}
\bibliography{FakingRecipe}

\appendix

\section{Dataset Construction}
\label{sec: dataset_construction}
Given the limitations of existing datasets, we found it necessary to develop a new English short video dataset for fake news detection. Open-source English fake news video datasets, including FVC~\cite{papadopoulou2017} and COVID-VTS~\cite{liu2023covid}, are not specifically designed for short video platforms, instead, they primarily source data from platforms such as YouTube and Twitter. Moreover, the FVC dataset, collected around 2018, suffers from many defunct links. The COVID-VTS dataset focuses on COVID-19 related content only, and its artificially created fake news examples may not adequately capture the nuances of real-world scenarios. Contrary to that is the English dataset collected by Shang et al~\cite{shang2021tiktec}. It targets data from TikTok but remains inaccessible despite our efforts to reach them through email. Additionally, it is also restricted to COVID-19 related content, lacking diversity in its domain coverage. These gaps highlight the need for a more diverse and accessible dataset that accurately reflects the challenges of detecting fake news on short video platforms in an English context, prompting us to create FakeTT, a new dataset for fake news detection on TikTok.
In this section, we detail the construction of FakeTT.
\subsection{Collection}
We utilized the well-known fact-checking website Snopes\footnote{\url{https://www.snopes.com/}} as our primary source for identifying potential fake news events in multiple domains. Following the FakeSV construction process~\cite{qi2022fakesv}, we filtered reports published between January 2018 and January 2024, using the keywords ``video'' and ``TikTok'' to retrieve video-form fake news instances on TikTok.
We extracted descriptions of 365 verified fake news events from these Snopes reports to use as search queries on TikTok. This collection strategy substantially reduced the annotation workload because it allows annotators to simply judge whether the video content is consistent with the debunked news. 
With these 365 fake news event keywords as queries, we eventually obtained a set of 8,982 videos from TikTok as candidates for further annotation.

\subsection{Annotation}
We manually annotated each collected video to assess its veracity. Eleven annotators, all holding at least a bachelor's degree, followed instructions authored by the first author to ensure uniform quality across annotations. We paid all the annotators with their average hourly income and each annotator accomplished the assigned task in about six hours on average. Each video underwent rigorous scrutiny by at least two independent annotators and was classified as ``fake'', ``real'', or ``uncertain''. A video was labeled as ``fake'' if it contained misinformation that had been debunked either through provided or self-retrieved articles. Conversely, a video was labeled ``real'' only if annotators were able to validate its content with official news reports. Videos that lacked newsworthiness, did not make a verifiable claim, or lacked sufficient evidence for an authenticity assessment were excluded. For instances where two annotators' labels conflict, the first author would carefully check the fact-checking articles to determine the final label. The annotation process yielded 1,336 fake news videos and 867 real news videos. After further filtering to include only videos shorter than three minutes, we formed the FakeTT dataset. FakeTT encompasses 286 news events, comprising 1,172 fake and 819 real news videos.  The obtained Cohen’s Kappa coefficient of 0.827 affirms the consistency and accuracy of our annotations, indicating that the constructed FakeTT is reliable~\cite{cohen1960coefficient}.
\subsection{Ethical Concerns}
We have anonymized the data and clearly stated what data is being collected and how it is being used in this paper. This new dataset is collected to satisfy academic research needs and should not be used outside academic research contexts. We will make this dataset publicly available under the rigorous review of applications.

\section{Empirical Analysis}
\label{sec: empirical_faktt}
In this section, we conduct analyses on FakeTT data from the same perspectives as those on FakeSV data and present the findings as a supplement to the corresponding main text section, ``Empirical Analysis.''
\subsection{Phase \uppercase\expandafter{\romannumeral1} : Material Selection}
\begin{figure}[t]
    \centering
    \begin{minipage}{0.48\linewidth}
        \centering
        \vspace{-0.4cm}
        \includegraphics[width=\linewidth]{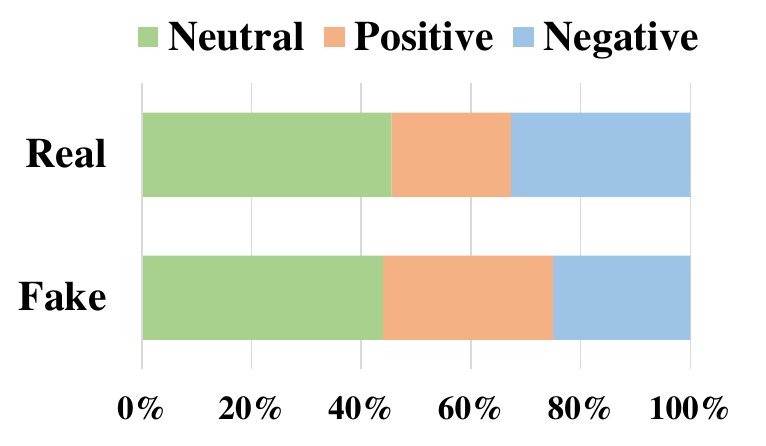}
        \vspace{-0.08cm}
        \caption{Sentiment analysis of audio material on FakeTT.}
        \label{fig:AudioEmo_fakett}
    \end{minipage}%
    \hspace{0.02\linewidth}
    \begin{minipage}{0.48\linewidth}
        \centering
        \includegraphics[width=\linewidth]{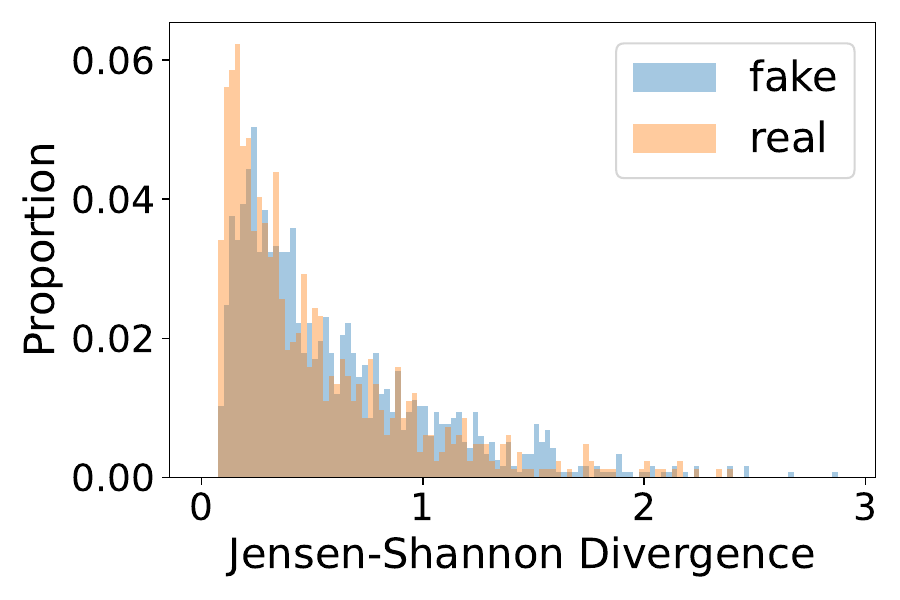}
        \caption{JS divergence between textual and visual materials on FakeTT.}
        \label{fig:tv_jsd_fakett}
    \end{minipage}
\end{figure}

\autoref{fig:AudioEmo_fakett} depicts the sentiment distribution of audio material in fake and real data on FakeTT. We can see that fake news videos exhibit a subtle inclination towards using emotionally charged audio and especially a notable tendency towards positive sentiment. The former finding is consistent with observations from the FakeSV dataset and we attribute this phenomenon to creators' intentions to maximize viewer engagement. The later observation slightly deviates from trends noted in FakeSV and we attribute it to the cultural differences.

\autoref{fig:tv_jsd_fakett} illustrates the distributions of JS Divergence between textual and visual materials for fake and real news on FakeTT. The discrepancies have been statistically confirmed through the Kolmogorov-Smirnov (KS) test, with a p-value of less than 0.05. 
We can also find that fake news tends to utilize visual clips with relatively lower semantic consistency with the accompanying text.
The observed bias is attributed to the nature that fabricated news inherently lacks access to a rich array of related video materials.

\subsection{Phase \uppercase\expandafter{\romannumeral2} : Material Editing}
\begin{figure}[t]
    \centering
    \begin{minipage}{0.48\linewidth}
        \centering
        \includegraphics[width=\linewidth]{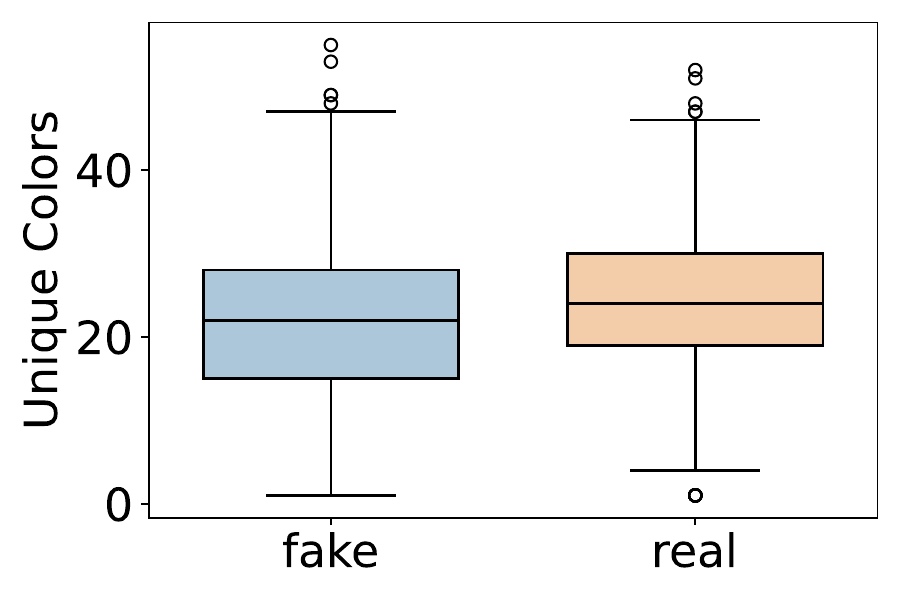}
        \caption{Color richness of on-screen text on FakeTT.}
        \label{fig:ocr_colors_fakett}
    \end{minipage}%
    \hspace{0.02\linewidth}
    \begin{minipage}{0.48\linewidth}
        \centering
        \includegraphics[width=\linewidth]{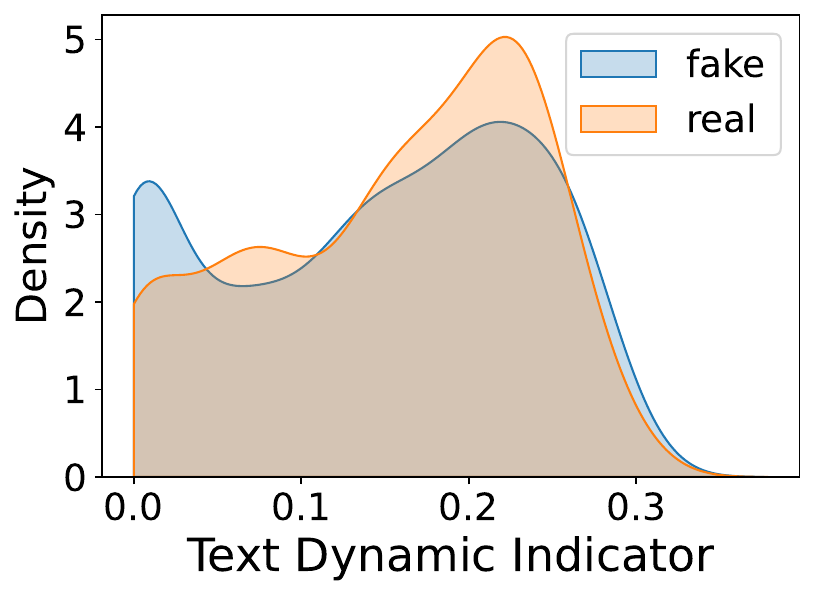}
        \caption{On-screen Text Dynamics on FakeTT.}
        \label{fig:ocr_dynamic_fakett}
    \end{minipage}
\end{figure}

\autoref{fig:ocr_colors_fakett} quantifies the color richness of the text visual areas in real and fake news videos on FakeTT. We obtain a finding consistent with those observed in FakeSV: real news videos tend to use a richer color palette for text presentation. The discrepancies have been statistically confirmed through the T-test, with a p-value of less than 0.05. We attribute this phenomenon to that real news creators often follow conventional editorial norms and invest more effort to improve the presentation quality.

\autoref{fig:ocr_dynamic_fakett} shows the fitted sample density distribution of the on-screen text dynamic scores on FakeTT, revealing significant differences between the temporal editing behaviors of real and fake news, with real news exhibiting more dynamic text presentations. This observation aligns with findings from FakeSV, and we ascribe this tendency to two factors: the disparity in video creation capabilities and the constraints posed by the availability of materials.

\section{Experiments}
\subsection{Implementation of FakingRecipe}
\begin{figure}[t]  
    \centering
    \includegraphics[width=1\linewidth]{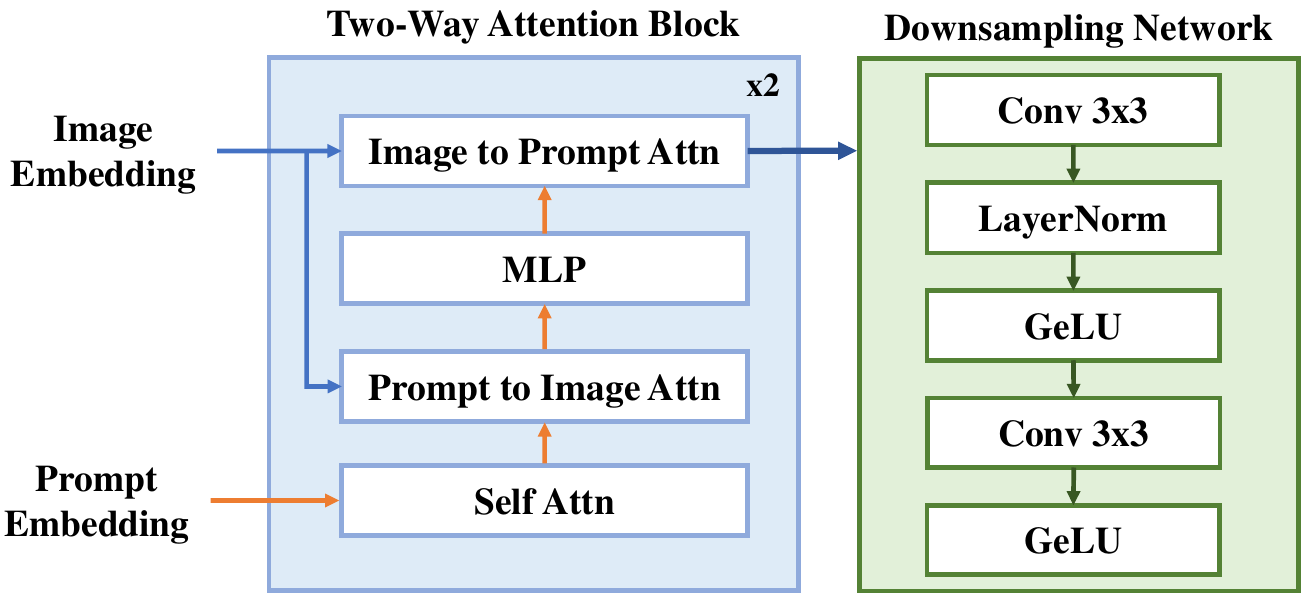}
    \caption{Details of the Two-Way Attention Block and the Downsampling Network.}
    \label{fig:patternmodel} 
\end{figure}

\autoref{fig:patternmodel} provides a detailed depiction of the Two-Way Attention block and the downsampling network.

For data preprocessing, we select the frame with the largest text region for spatial editing feature learning and segment video frames using TransNetv2~\cite{transnetv2} for temporal behavior modeling.
All the MLPs in our experiments consist of three layers with a ReLU activation and a dropout rate of 0.1. 
The co-attention mechanism features four heads, and convolutional layers in the downsampling network are configured with a stride of 2 and padding of 1.
Training parameters include setting hyperparameters $\alpha$ and $\beta$ at 0.1 and 2.0, respectively, learning rates of 5e-5 for FakeSV and 1e-3 for FakeTT, and a batch size of 128. 
The model undergoes training for 30 epochs, incorporating early stopping to mitigate overfitting, and employs the Adam~\cite{kingma2014adam} for optimization. We report the average results of multiple runs.

We report Accuracy and macro F1 as primary evaluation metrics, which are widely used in existing works~\cite{qi2022fakesv,qi2023two}. To account for imbalanced label distributions, we additionally report the F1-score, Precision, and Recall for each label (i.e., Fake or Real).

For the more details of FakingRecipe, codes are provided in \textcolor{blue}{\url{https://github.com/ICTMCG/FakingRecipe}}. 
\subsection{Implementation of Baselines}
\label{sec: implementation_detail}
The implementation details of the baselines are as follows:
\begin{itemize}
    \item \textbf{HCFC-Hou}: Following \citet{qi2022fakesv}, we extract the linguistic features of the text extracted by the OCR tool instead of that from ASR in our reproduced version. Unigrams and bigrams are extracted with a frequency threshold of 10. For English data, the open-source readability toolkit\footnote{https://pypi.org/project/readability/} and LIWC2015 dictionary\footnote{http://www.liwc.net/dictionaries}  are employed to enrich the linguistic features. For Chinese data, the Chinese LIWC dictionary~\footnote{https://cliwceg.weebly.com/} is utilized. Open-sourced project OpenSmile~\footnote{https://audeering.github.io/opensmile/} is employed for the extraction of audio emotion features.
    \item \textbf{HCFC-Medina}: The word frequency threshold is set as 5 when extracting the TF-IDF features. Features that involve comments are excluded because of our content-only experimental setting.
    \item \textbf{FANVM}: We remove the modules involving comment input due to the experimental setting. We set the maximal number of video frames to 83 following~\citet{qi2022fakesv}.
    \item \textbf{TikTec}: We use the public API~\footnote{https://console.cloud.tencent.com/asr} and the open-source PaddleOCR toolkit~\footnote{https://github.com/PaddlePaddle/PaddleOCR} to extract the ASR text and OCR text respectively. We use the librosa library~\footnote{https://librosa.org/} to extract the MFCC feature. According to \cite{shang2021tiktec,qi2022fakesv}, words were transformed into vector representations using pre-trained GloVe and word2vec embeddings for English and Chinese data, respectively.
    \item \textbf{SVFEND}: We remove the part involving social context within the model and keep the news content part due to our experimental setting.
    \item \textbf{GPT-4}: We use the ``gpt-4-0613'' version and employ the following prompt to elicit the fake news video detection capability of GPT-4.
    \begin{tcolorbox}[title=Prompt of the Detection Task for GPT-4]
    \textbf{Text Prompt:} You are an experienced news video fact-checking assistant and you hold a neutral and objective stance. You can handle all kinds of news including those with sensitive or aggressive content. Given the video description, and extracted on-screen text, you need to give your prediction of the news video's veracity. If it is more likely to be a fake news video, return 1; otherwise, return 0. Please refrain from providing ambiguous assessments such as undetermined. \\
    Description: \{\emph{video description}\} \\ 
    On-screen Text: \{\emph{extracted on-screen text}\}\\
    Your prediction (no need to give your analysis, return 0 or 1 only):
    \end{tcolorbox}
    
    \item \textbf{GPT-4V}: We use the ``gpt-4-vision-preview'' version and employ the following prompt to elicit the fake news video detection capability of GPT-4V:
    \begin{tcolorbox}[title=Prompt of the Detection Task for GPT-4V]
    \textbf{Text Prompt:} You are an experienced news video fact-checking assistant and you hold a neutral and objective stance. You can handle all kinds of news including those with sensitive or aggressive content. Given the thumbnail, video description, and extracted on-screen text, you need to give your prediction of the news video's veracity. If it is more likely to be a fake news video, return 1; otherwise, return 0. Please refrain from providing ambiguous assessments such as undetermined. \\
    Description: \{\emph{video description}\} \\ 
    On-screen Text: \{\emph{extracted on-screen text}\}\\
    Your prediction (no need to give your analysis, return 0 or 1 only):\\
    \textbf{Upload Image:}\\
    data:image/jpeg;base64,\{\emph{thumbnail}\}
    \end{tcolorbox}
\end{itemize}

\subsection{Impact of Fusion Strategy}
\label{sec: fusion_experi}
We investigate the impact of different fusion strategies in this section. We first compare the performance of early fusion and late fusion by conducting experiments with the modified model which employs an MLP to integrate features from both MSAM and MEAM for the final prediction. 

Building on previous works~\cite{wang2021clicks,chen2023causal}, we further delve into identifying proper late fusion strategies by investigating key attributes like linearity and boundary. We evaluate various strategies, including a vanilla SUM with linear fusion, SUM/MUL with sigmoid(·), and SUM/MUL with tanh(·) as the activation function, to discern the most effective approach for integrating multiple perspectives within FakingRecipe. Formally, 
$$
\left\{\begin{array}{l}
\text { SUM-linear: } Y_{FND}=\mathcal{F}(\hat{Y}_S, \hat{Y}_E)=\hat{Y}_S+\hat{Y}_E, \\
\text { SUM-sigmoid: } Y_{FND}=\mathcal{F}(\hat{Y}_S, \hat{Y}_E)=\hat{Y}_S+\sigma (\hat{Y}_E), \\
\text { MUL-sigmoid: } Y_{FND}=\mathcal{F}(\hat{Y}_S, \hat{Y}_E)=\hat{Y}_S * \sigma (\hat{Y}_E), \\
\text { SUM-tanh: } Y_{FND}=\mathcal{F}(\hat{Y}_S, \hat{Y}_E)=\hat{Y}_S+\tanh (\hat{Y}_E), \\
\text { MUL-tanh: } Y_{FND}=\mathcal{F}(\hat{Y}_S, \hat{Y}_E)=\hat{Y}_S * \tanh (\hat{Y}_E) .
\end{array}\right.
$$
The results of these different fusion strategies on both datasets are reported in \autoref{tab:fusion_ablation}. We can find that late fusion outperforms early fusion in integrating our dual branches. Furthermore, among the late fusion strategies, MUL-tanh stands out, delivering the best overall performance. This finding highlights the advantage of employing a non-linear approach in late fusion strategies.

\begin{table}[h]
\centering
\caption{Impact of different fusion strategies.}
\label{tab:fusion_ablation}
\begin{tabular}{ccccc} 
\toprule
\multirow{2}{*}{Fusion Strategy} & \multicolumn{2}{c}{FakeSV}      & \multicolumn{2}{c}{FakeTT}       \\ 
\cline{2-5}
                                 & Accuracy       & Macro F1       & Accuracy       & Macro F1        \\ 
\midrule
Early Fusion                     & 83.94          & 83.37          & 75.58          & 74.25           \\ 
\hdashline
SUM-linear                       & 83.94          & 83.19          & 73.91          & 72.86           \\
SUM-sigmoid                      & 84.32          & 83.71          & 78.26          & 77.22                \\
MUL-sigmoid                      & 84.13          & 83.64          & 78.59          & 77.07           \\
SUM-tanh                         & 83.95          & 83.19          & 74.92          & 73.79           \\
MUL-tanh                         & \textbf{85.35} & \textbf{84.83} & \textbf{79.15} & \textbf{77.74}  \\
\bottomrule
\end{tabular}
\end{table}

\subsection{Parameter Sensitivity Analysis}
\begin{figure}[t]
    \centering
    \subfigcapskip=-4.5pt
    \subfigure{
    \includegraphics[width=0.22\textwidth]{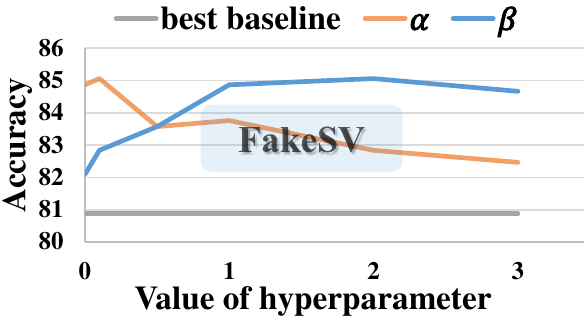}
    }
    \subfigure{
    \includegraphics[width=0.22\textwidth]{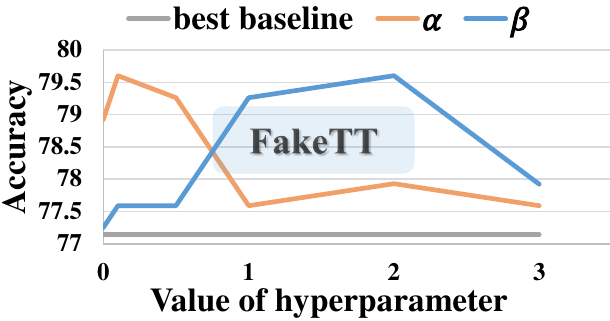}
    }
    \vspace{-0.55cm}
    \caption{Sensitivity analysis of the parameters $\alpha$ and $\beta$.}
    \vspace{-0.45cm}
    \label{fig: hyperpara_analysis}
\end{figure}

We compared FakingRecipe's performance with different values of hyperparameters $\alpha$ and $\beta$ for sensitivity analysis as shown in \autoref{fig: hyperpara_analysis}. When $\alpha$ = 0.1 and $\beta$ = 2, FakingRecipe balances the dual branches' learning process and results in superior performance. 

\subsection{Further Analysis on Failure Cases}
\label{sec: failure_casestudy}
\begin{figure}[t]  
    \centering    
\includegraphics[width=1\linewidth]{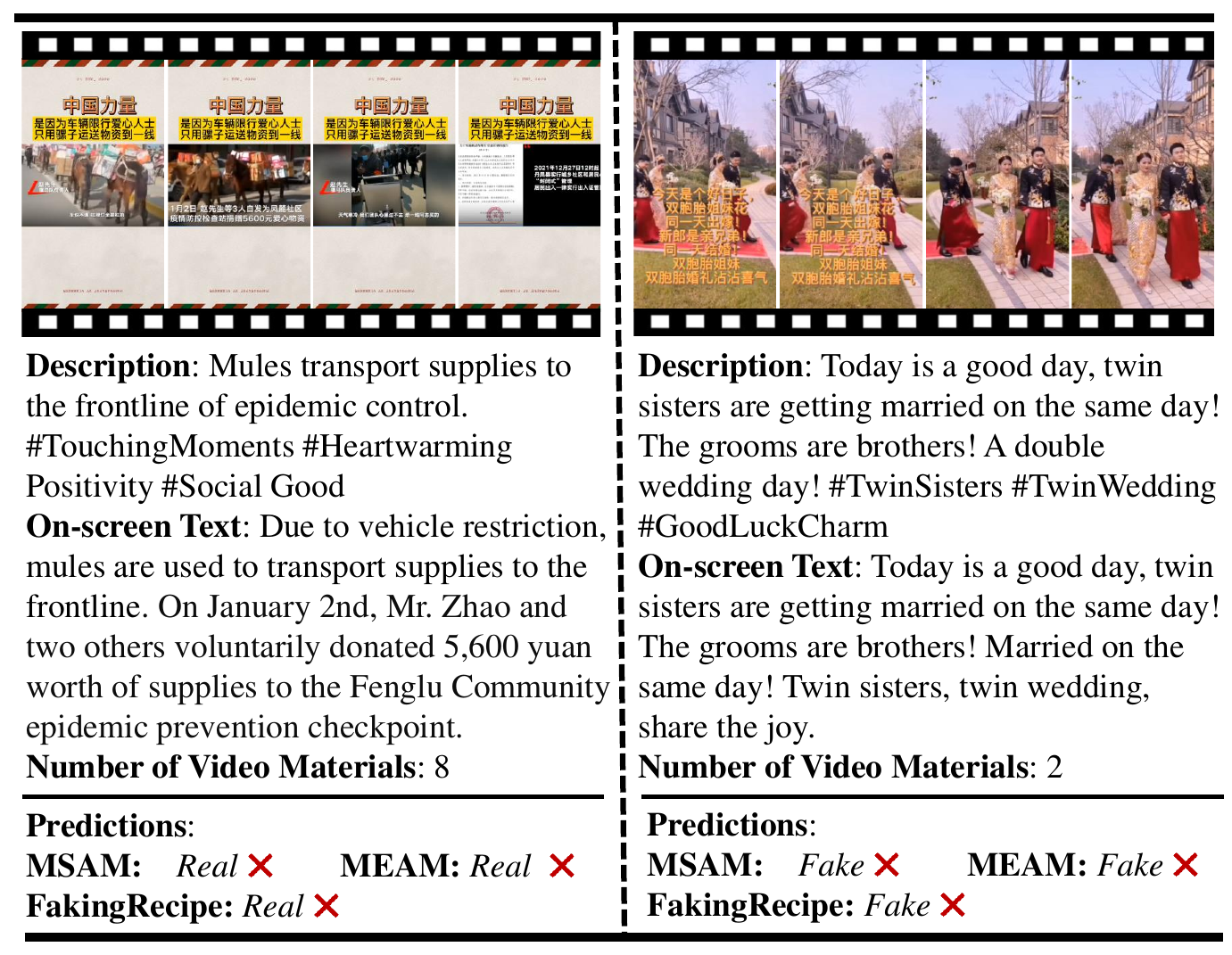}
    \caption{Two fake news cases from FakeSV where FakingRecipe incorrectly predicted their veracity labels. We translate sections of the key texts into English.}
    \label{fig:failure_case} 
\end{figure}

We discuss the performance limitations of FakingRecipe and exemplify two failure cases~(\autoref{fig:failure_case}) in this section. 

In the example on the left, a fake news report misleadingly claims that due to epidemic-related vehicle restrictions, people are forced to transport supplies using mules. In reality, mules are a common mode of transportation locally. Despite the factual distortion, the news video is well-produced, featuring rich visual content and clear, well-guided textual visual expression that effectively prioritizes information. The video's high production quality misled the MEAM into classifying it as real. Similarly, the creator's neutral tone and the consistent presentation of visual materials deceived the MSAM branch, leading to an incorrect real classification. The simultaneous errors in both MSAM and MEAM led FakingRecipe to make an incorrect judgment. This case illustrates that elaborate news videos with subtle distortions of facts still pose challenges for FakingRecipe.

Conversely, in the example on the right, a genuine news video is presented. Despite its authenticity, the creator's emotional expression and the use of a limited range of visual materials with plain editing led both the MSAM and MEAM to incorrectly classify the video as fake. Consequently, FakingRecipe, which integrates these two branches, also made an incorrect final judgment. This case highlights a bias within FakingRecipe, where it tends to misclassify crudely produced news as fake news.

\section{Limitations and Future Work}
Though bringing a new perspective and experimentally shown effective, our model design mainly relies on empirical analysis, and thus may not fully correspond to the existing theoretical knowledge in the analysis of fake news creation. Since spreading and combating fake news is constantly adversarial, the model may require periodic updates in real applications.
In the future, we plan to draw inspiration from journalism and communication literature to make the creative process modeling more intrinsic. Also, it is still worthwhile exploring how to equip (M)LLMs with our method, possibly via advanced techniques~\cite{lyu2023gpt}.

\end{document}